%% file: aaai2026.tex
\documentclass[letterpaper]{article} 
\usepackage{aaai2026}  
\usepackage{times}  
\usepackage{helvet}  
\usepackage{courier}  
\usepackage[hyphens]{url}  
\usepackage{graphicx} 
\usepackage{natbib}        
\usepackage{caption}       
\usepackage[T1]{fontenc}   
\usepackage{url}           
\usepackage{booktabs}      
\usepackage{amsfonts}      
\usepackage{amssymb}       
\usepackage{amsmath}       

\usepackage{nicefrac}      
\usepackage{microtype}     
\usepackage{multirow}
\usepackage{datetime2}
\usepackage{color}
\usepackage{mathrsfs}
\usepackage{subcaption}
\usepackage{graphicx}
\usepackage{adjustbox}
\usepackage{soul}
\usepackage{times}
\usepackage{helvet}
\usepackage{courier}
\usepackage{xcolor}
\usepackage{listings}
\usepackage{mathtools} 
\usepackage{enumitem}
\usepackage{xcolor}
\usepackage{arydshln}
\usepackage{algorithm}      
\usepackage{algorithmicx}   
\usepackage{algpseudocode}  
\usepackage{amsthm}                      
\newtheorem{definition}{Definition}
\usepackage{amsthm}
\usepackage{makecell}
\newtheorem{Lemma}{Lemma}
\usepackage{mdframed}      
\newmdenv[linecolor=black,
          leftline=true,
          topline=false,
          bottomline=false,
          rightline=false,
          skipabove=\baselineskip,
          skipbelow=\baselineskip]
         {mybar}
\usepackage[dvipsnames]{xcolor}
\usepackage{pifont}
\usepackage{textcomp}
\usepackage{xspace}               
\newcommand{\ours}{\textbf{FiD-GP}\xspace}
\usepackage{newfloat}      
\DeclareCaptionStyle{ruled}{labelfont=normalfont,labelsep=colon,strut=off} 
\floatstyle{ruled}
\newfloat{listing}{tb}{lst}{}
\floatname{listing}{Listing}

\input{macros/macros.tex}

\title{Flow‑Induced Diagonal Gaussian Processes}
\author {
    Moule Lin\textsuperscript{\rm 1},
    Andrea Patane\textsuperscript{\rm 1},
    Weipeng Jing\textsuperscript{\rm 2},
    Shuhao Guan\textsuperscript{\rm 3},
    Goetz Botterweck\textsuperscript{\rm 1},
}
\affiliations {
    \textsuperscript{\rm 1} Trinity College Dublin, Dublin, Ireland\\
    \textsuperscript{\rm 2} Northeast Forestry University, Heilongjiang, China\\
    \textsuperscript{\rm 3} University College Dublin,  Dublin, Ireland\\
    \{moulel, apatane, goetz.botterweck\}@tcd.ie, jwp@nefu.edu.cn, shuhao.guan@ucdconnect.ie
 
}

\usepackage{bibentry}

\begin{document}

\maketitle

\begin{abstract}
We present Flow-Induced Diagonal Gaussian Processes (\ours), a compression framework that incorporates a compact inducing weight matrix to project a neural network's weight uncertainty into a lower-dimensional subspace. Critically, \ours relies on
normalising-flow variational posterior and spectral regularisations to augment its expressiveness and align the inducing subspace with feature-gradient geometry through a numerically stable projection mechanism objective. Furthermore, we demonstrate how the prediction framework in \ours can help to design a single-pass projection for Out-of-Distribution (OoD) detection. 
%
Our analysis shows that \ours improves uncertainty estimation ability on various tasks compared with SVGP-based baselines, satisfies tight spectral-residual bounds with theoretically guaranteed OoD detection, and significantly compresses the neural network's storage requirements at the cost of increased inference computation dependent on the number of inducing weights employed. 
%
Specifically, in a comprehensive empirical study spanning regression, image classification, semantic segmentation, and out-of-distribution detection benchmarks, it cuts Bayesian training cost by several orders of magnitude, compresses parameters by roughly \(51\%\), reduces model size by about \(75\%\), and matches state-of-the-art accuracy and uncertainty estimation.

\textbf{Code}: \url{https://github.com/anonymouspaper987/FiD-GP.git}
\end{abstract}

\input{section/1-introduction}

\input{section/2-related}

\input{section/3-varitional}
\input{section/4-method}

\input{section/5-experiments}

\input{section/6-discussion}

\input{section/6.5-conclusion}

\bibliography{aaai2026}
\input{section/8-checklist}

\input{section/6-appendix}

\end{document}

%% file: macros/macros.tex
\definecolor{gbcolor}{rgb}{0.8,0,0.8}
\definecolor{apcolor}{rgb}{0,0.7,0.2}
\definecolor{dpcolor}{rgb}{0,0.8,0}

\makeatletter
\newcommand{\del@strikeout}[1]{{\color{red}\sout{#1}}} 
\newcommand{\del@remove}[1]{}                          
\newcommand{\del@show}[1]{#1}                          

\newcommand{\setdelmode}[1]{%
  \@ifundefined{del@#1}{%
    \PackageWarning{DeleteMacro}{Unknown mode '#1'. Using 'strikeout'.}%
    \let\del\del@strikeout
  }{%
    \expandafter\let\expandafter\del\csname del@#1\endcsname
  }%
}
\makeatother

\setdelmode{strikeout}


%% file: section/1-introduction.tex
\section{Introduction}
Reliable uncertainty estimates are especially crucial and sought after in safety-critical applications of neural networks, like autonomous driving \citep{hubmann2017decision}, medical diagnosis \citep{chua2023tackling}, and many others \citep{blasco2024survey,hernandez2015probabilistic,zyphur2015bayesian}.
Research on predictive uncertainty has expanded in multiple directions. Bayesian Neural Networks (BNNs) \citep{kononenko1989bayesian,mackay1995bayesian,thodberg1996review}, ensemble methods \citep{hoffmann2021uncertainty,rahaman2021uncertainty}, and distance-aware frameworks \citep{mukhoti2023deep,liu2020simple,zhang2024discriminant} have emerged as prominent approaches that produce strong performance on estimating uncertainty and related benchmarks. In these settings, models are expected not only to deliver accurate predictions but also to generate reliable confidence estimates, particularly for identifying Out-of-Distribution (OoD) inputs.

\begin{figure}[t]
\centering
\includegraphics[width=0.26\textwidth]{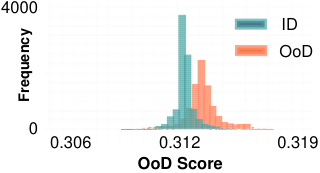}
\caption{Distribution of predictive scores generated by the ResNet-18 model equipped with Sparse Variational Gaussian Processes (SVGP). In-distribution (ID) dataset is CIFAR-100, Out-of-Distribution (OoD) dataset is CIFAR-10.}
\label{fig:idood}
\end{figure} 
Unfortunately, although progress has been made, the more significant computational overhead required compared to deterministic counterparts during training and inference still limits the widespread adoption of these approaches in practical industry settings. 
In turn, it has spurred research into more efficient and effective uncertainty estimation methods, such as Rank-1 BNN \citep{dusenberry2020efficient}, lower uncertainty space (Sparse Gaussian Processes or VAE) \citep{ritter2021sparse, franchi2023encoding}, distance-aware frameworks \citep{van2020uncertainty, liu2020simple,mukhoti2023deep} as well as quantisation methods \citep{lin2023quantization,lin2025stochastic, hubin2024sparse,ritter2021sparse}.

Meanwhile, Gaussian Processes (GP) \citep{seeger2004gaussian,williams1995gaussian,hida1993gaussian,opper2000gaussian} can represent an infinitely wide network with a finite number of parameters. 
Sparse Gaussian Processes (SGPs) \citep{snelson2005sparse,kuss2005assessing,wei2024scalable} project the full GP onto a small inducing matrix and thereby dramatically lower computational cost by restricting uncertainty to a low-dimensional subspace. 
%
%
%
Unfortunately, despite their elegance, sparse‑GP wrappers on BNNs still underperform on deep networks as limited expressiveness, stemming from their simple Gaussian variational distribution over a low‑dimensional inducing subspace, restrictive stationary kernel assumptions, and many other factors, not only degrades predictive accuracy and uncertainty calibration \citep{swiler2020survey, lawrence2002fast} but also fails to separate In‑distribution from Out-of-Distribution input.

As shown in Figure~\ref{fig:idood} for example, the predictive‐score distributions produced by the ResNet‑18+SVGP model on CIFAR‑100 (ID) and CIFAR‑10 (OoD) overlap substantially, indicating that the simple Gaussian variational distribution cannot adequately discriminate OoD samples.

In this paper, we present a framework that shapes the SGP inducing variational distribution with a normalising flow equipped with spectral regularisation to model complex, multi‑modal feature correlations. We adapted a Kronecker‑structure to present the covariance of the inducing matrix and its associated weights for efficient training and inference.
In the context of uncertainty's practical applications, we further showcase how our modelling framework can be used to derive an Out-of-Distribution (OoD) detection system based on a process that projects the feature space into the inducing‑matrix space, with theoretical guarantees for feature–inducing alignment provided by the normalising flow and spectral regularisation, which forms a \textbf{single-pass ID/OoD} detection mechanism.

We perform an extensive empirical investigation on the effectiveness of our method across regression, image classification, and semantic segmentation tasks. We utilise a synthetic 1-D function for evaluating regression performance, following \citep{miyato2018spectral} and \citep{ritter2021sparse}, and conduct classification experiments on CIFAR-100~\citep{krizhevsky2009learning} and ImageNet-1k~\citep{deng2009imagenet} using ResNet-18~\citep{he2016deep} as the base backbone. For semantic segmentation, we validate our method on widely adopted benchmarks, including CamVID~\citep{brostow2009semantic} and CityScapes~\citep{cordts2016cityscapes}, using FCN-ResNet50~\citep{long2015fully} and HRNet-W48~\citep{sun2019high} as the backbone networks. Our approach consistently achieves state-of-the-art performance compared with several recent methods, including strong deterministic baselines, BatchEnsemble, FFG‑U \citep{ritter2021sparse}, and F‑SGVB‑LRT \citep{nguyen2024flat}.

\par
\textbf{Our main contributions:}
\par
\begin{itemize}
  \item We propose Flow-Induced Diagonal Gaussian Processes (\ours), a GP-inspired uncertainty module that integrates with off-the-shelf BNN architectures and builds on sparse Gaussian processes, placing a normalizing-flow on variational posterior distribution, while the GP prior remains Gaussian and the standard prior-conditional $p(W\mid u)$ is preserved.
  
  \item We develop a jittered-Cholesky projection objective that aligns inducing-point subspaces with feature-gradient geometry, producing a single-pass projection score with tight spectral-residual bounds and theoretically guaranteed near-perfect OoD discrimination.
  \item We conducted comprehensive empirical experiments across regression, image classification, semantic segmentation, and Out-of-Distribution detection benchmarks, demonstrating significant reductions in training cost, approximately 51\% parameter compression, and achieving state‑of‑the‑art accuracy and uncertainty estimation without heavy post‑hoc calibration.
\end{itemize}

%% file: section/2-related.tex
\section{Related Work}
Uncertainty estimation for modern artificial neural networks remains a long-standing and significant challenge, particularly in safety-critical domains \citep{blei2017variational,snelson2007local,arhonditsis2017uncertainty}. Many studies have investigated the balance of efficiency and expressiveness in estimating uncertainty from different perspectives. In this work we look at combining GPs, in particular sparse GPs, together with BNNs.
\par
%
Considering the inference efficiency, several works have looked at decomposing the inducing points or redesigning the inference process. Decoupled Gaussian Processes (DGPs) \citep{cheng2015learn} push SVGPs, separating the bases for the mean and covariance, resulting in the mean growing linearly without enlarging the cubic bottleneck. Greater expressiveness can also be achieved through structured inducing domains. 
This was later extended to Convolution \citep{van2017convolutional} through the construction of an inter-domain inducing matrix approximation that is well-tailored to the convolutional kernel. 
%
Recent research has proposed Gaussian posterior approximations for BNNs with efficient covariance structures \cite{ritter2018scalable,mishkin2018slang}. 
\par
SVGPs scale exact GPs to $\mathcal O(M^3)$ via $M\ll N$ inducing points \citep{titsias2009variational,hensman2015mcmc}, where $N$ is the total number of parameters of the neural network and $M$ denotes the dimensionality of the inducing matrix, but their Gaussian variational posterior is limited in expressivity \citep{titsias2009variational}. To enrich this, normalising flows, e.g.\ Real NVP \citep{dinh2017density}, MAF \citep{papamakarios2017masked}, FFJORD \citep{grathwohl2018ffjord} and Inverse Autoregressive Flow \citep{kingma2016improved} have been applied to the inducing points outputs to capture more flexible, non-Gaussian marginals and tighter ELBOs \citep{rezende2015variational,cutajar2019preconditioning,rezende2015variational,kingma2016improved}.

\par
Beyond modelling predictive uncertainty within the training distribution, several works seek to detect and properly score samples that fall outside it. The earliest attempts rely on likelihood-based generative models, where normalising flows or autoregressive densities assume higher log-likelihood on in-distribution data than OoD inputs \citep{dinh2017density,kingma2018glow,nalisnick2019do,ren2019likelihood}. This further motivates the reconstruction-error paradigm, which treats the difficulty of rebuilding an input via autoencoders or memory-augmented networks as an anomaly signal \citep{an2015variational,gong2019memorizing}. 
%
Recently, there has been increasing interest in integrating the Gaussian Process (GP) perspective into deep architectures: distance-aware single-pass heads (SNGP) \citep{liu2020simple}, kernel-density hybrids (DUQ/DUE) \citep{van2020uncertainty}, deterministic GP post-processing (DDU) \citep{mukhoti2023deep}, logit-level GP unification \citep{chen2024uncertainty}, and sparse or geometric variants.

These works investigated the trade-off between efficiency and expressiveness; however, they still either sacrifice closed‑form calibration, rely on multiple forward passes, or impose specialised architectural constraints. Instead, our approach shares parameters
via the flow-augmented variational posterior with an efficient low-rank structure.

%% file: section/3-varitional.tex
\section{Preliminaries}
\label{preliminary}
\subsection{Sparse Gaussian Process}
We first reviews the core concepts of SGPs and their variational inference, then show how to integrate them with BNNs to enable end-to-end uncertainty estimation.

A Gaussian Process (GP) \citep{williams1995gaussian} defines a distribution over functions:
$
f(\cdot)\sim\mathcal{GP}\bigl(m(\cdot),\,\Sigma(\cdot,\cdot)\bigr)
$
where $m$ is the mean and $\Sigma(\cdot,\cdot)$ the kernel.

Introduce $M\ll N$ inducing points $\mathbf U$ with variables $\mathbf u=f(\mathbf U)$.  
We place the prior
\begin{equation}\label{eq:u_prior_sigma}
p(\mathbf u)=\mathcal{N}\bigl(\mathbf 0,\,\Sigma_{MM}\bigr)
\end{equation}
and approximate the posterior by
\begin{equation}\label{eq:elbo_sigma}
\mathcal{L}
=\sum_{n=1}^N\mathbb{E}_{q(f_n)}\bigl[\log p(y_n\mid f_n)\bigr]
-\mathrm{KL}\bigl[q(\mathbf u)\,\|\,p(\mathbf u)\bigr]
\end{equation}
The conditional prior over \(\mathbf f=[f(\mathbf x_n)]_{n=1}^N\) is
{\small
\begin{equation}
\label{eq:condition}
p(\mathbf w\mid\mathbf u)
=\mathcal{N}\Bigl(\Sigma_{NM}\,\Sigma_{MM}^{-1}\mathbf u,\;
\Sigma_{NN}-\Sigma_{NM}\,\Sigma_{MM}^{-1}\,\Sigma_{MN}\Bigr)
\end{equation}
}
where $\Sigma_{NM}[n,m]=K(\mathbf x_n,\mathbf z_m)$ and $\Sigma_{NN}[n,n']=K(\mathbf x_n,\mathbf x_{n'})$. $K$ is the kernel function, and here we identify $\mathbf w=\mathbf f$ as the vector of weights in our Bayesian Neural Network.  
This reduces inference to $\mathcal{O}(NM^2+M^3)$ time and $\mathcal{O}(NM+M^2)$ memory.

(For more details see Appendix A.)

\subsection{Kronecker‐Structured Covariance}
\label{sec:Kronecker}
Using Kronecker identities, the transforms become
\begin{equation}\label{eq:row_col_structure_refined}
T_{\mathrm{row}}
=\Sigma_{W,U}^{(\mathrm{row})}\bigl(\Sigma_{U}^{(\mathrm{row})}\bigr)^{-1},
\quad
T_{\mathrm{col}}
=\Sigma_{W,U}^{(\mathrm{col})}\bigl(\Sigma_{U}^{(\mathrm{col})}\bigr)^{-1}
\end{equation}
Hence the conditional mean of \(W\) given \(U\) is simply
\begin{equation}\label{eq:w_refined}
\mathbb{E}[W\mid U]
= T_{\mathrm{row}}\;U\;T_{\mathrm{col}}^\top
\end{equation}
and sampling follows Matheron’s rule:
\begin{equation}\label{eq:matheron_refined}
W\mid U
= W_{\mathrm{prior}}
+T_{\mathrm{row}}\bigl(U-U_{\mathrm{prior}}\bigr)T_{\mathrm{col}}^\top
\end{equation}
Here $\Sigma_{WU}$ and $\Sigma_{UU}$ denote \emph{prior} covariance blocks (possibly with Kronecker factorization);
they do not depend on the variational posterior $q(u)$.
(The detailed derivation is provided in Appendix B.)

\noindent\textbf{Whitened Representation:}  
To improve numerical stability during sampling, we adopt a whitened parameterization. Let $K_U = L L^\top$ be the Cholesky decomposition of the prior covariance, and define the whitened variable $\mathbf{v} = L^{-1}\mathbf{u}$. The prior becomes:
$
p(\mathbf{v}) = \mathcal{N}(\mathbf{v} \mid \mathbf{0}, I_M)
$
with variational distribution $q(\mathbf{v}) = \mathcal{N}(\mathbf{v} \mid \tilde{\mathbf{m}}, \tilde{\mathbf{S}})$. The Matheron sampling rule in whitened space is:
\begin{equation}
W \mid \mathbf{v} = W_{\mathrm{prior}} + T_{\mathrm{row}} L (\mathbf{v} - \mathbf{v}_{\mathrm{prior}}) T_{\mathrm{col}}^\top
\end{equation}
\subsection{Normalisation Flow with Spectral regularisation}
\label{spectral}
Spectral normalisation \cite{miyato2018spectral} enforces a 1‑Lipschitz constraint by normalising each weight matrix:
\begin{equation}
\tilde W = \frac{W}{\sigma_{\max}(W)},
\quad
\sigma_{\max}(W)=\sup_{\|x\|_2=1}\|W x\|_2.
\end{equation}
In a Normalising Flow $g_{\phi}$, each layer’s Jacobian $J_l=D_l\tilde W_l$ satisfies $|\det\tilde W_l|\le1$, so
\begin{equation}
\label{eq:jacobian}
|\det J_g|=\prod_l|\det J_l|\le\prod_{l,i}\phi_l'(h_{l,i})
\end{equation}
ensuring the overall Jacobian determinant remains tractable.

%% file: section/4-method.tex
\section{Methodology}
\begin{figure}[H]
\centering
\includegraphics[width=0.49\textwidth]{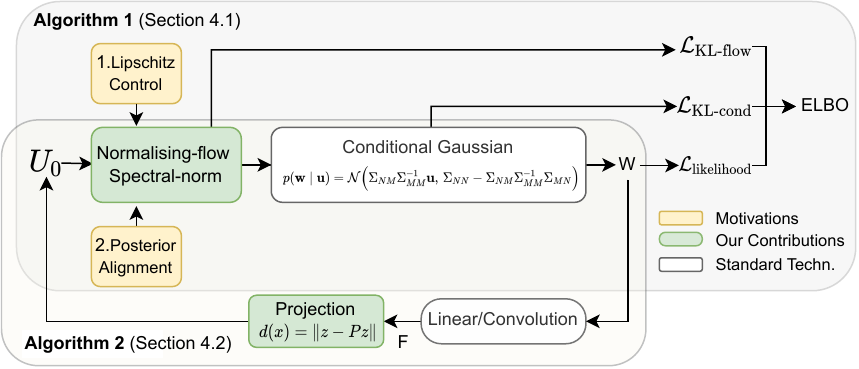}
\caption{Overview of our approach (\ours): Flow-based conditional Gaussian with spectral control and projection residual.}
\label{fig:overall}
\end{figure}

In this section, we divide our approach into two parts. The overall framework is shown in Figure \ref{fig:overall}, which comprises the standard uncertainty‐estimation model (Section 4.1) and a robust Out-of-Distribution detection module (Section 4.2).
%

\subsection{Flow-Based Variational Posterior for Uncertainty Estimation}

We initially apply a Normalising Flow with Spectral Normalisation to our inducing-point matrix (see Definition~\ref{Normalising}), so that the \emph{variational posterior} over inducing variables departs from a simple zero-mean Gaussian and can represent richer, non-Gaussian structure. Importantly, we do \emph{not} claim that the posterior over the inducing matrix is Gaussian; rather, by exploiting the conditional-Gaussian identity (Eq.~\ref{eq:condition}), the \emph{conditional} distribution $p(W\mid u)$ retains the standard GP Gaussian form for any $u$, while the marginal over weights $q(W)=\int p(W\mid u)\,q(u)\,du$ is generally non-Gaussian and is estimated via reparameterised Monte Carlo (see Appendix~B for details). Spectral normalisation enforces a 1-Lipschitz mapping that stables training and tractable Jacobian determinants in the flow-based variational posterior. 

\begin{definition}[Normalising Flow with Spectral Normalisation for the \textbf{Variational Posterior}]
\label{Normalising}
Let $q_0(z)=\mathcal{N}(z\mid 0,I_M)$ be a base Gaussian in $\mathbb{R}^M$, and let
$g_{\phi}:\mathbb{R}^M\to\mathbb{R}^M$ be a smooth bijection.
Defining $u=g_{\phi}(z)$, the variational posterior density is
\begin{equation}
q(u) \;=\; q_0\!\big(g_{\phi}^{-1}(u)\big)\,
\left|\det \nabla_{u}\,g_{\phi}^{-1}(u)\right|.
\end{equation}
\end{definition}

\par
Substituting the flow-based \(q(u)\) into standard SVGP Evidence Lower Bound (ELBO) function, and then the ELBO becomes: 
\begin{equation}
\label{eq:elbo}
\begin{aligned}
&\mathcal{L}_{\text{ELBO}}
= \underbrace{\mathbb{E}_{q_0,\epsilon}[\log p(\mathcal{D}\mid W)]}_{\text{Expected log-likelihood}} \\
&\quad -\underbrace{\mathbb{E}_{q_0}\left[\log q_0 - \log|\det J_g| - \log p(g(u_0))\right]}_{\text{KL divergence of flow-based inducing posterior}} \\
&\quad -\underbrace{\tfrac{D}{2}(\lambda^2 - 1 - 2\log\lambda)}_{\text{Conditional Gaussian KL}}
\\
&\text{(For more details see Appendix C.)}
\end{aligned}
\end{equation}
where $q_0$ is the base distribution of latent variables $u_0$, $\epsilon$ is Gaussian noise for reparameterisation, $W$ denotes neural network weights sampled from $u_0$ and $\epsilon$, $\mathcal{D}$ is the dataset, $g$ is a flow transformation with Jacobian determinant $|\det J_g|$, $p$ denotes prior or likelihood densities, $D$ is the weight dimension, and $\lambda$ controls the conditional Gaussian variance. 
\begin{table*}[!t]
\centering
\renewcommand{\arraystretch}{1.5}
\begin{tabular}{lcc}
\toprule
\toprule
\textbf{Method} & \textbf{Time complexity} & \textbf{Storage complexity}\\
\midrule
Deterministic & $\mathcal{O}(Nd_{in}d_{out})$ & $\mathcal{O}(d_{in}d_{out})$ \\
BatchEnsemble & $\mathcal{O}(NKd_{in}d_{out})$ & $\mathcal{O}(Kd_{in}d_{out})$ \\
FFG-$\mathbf{U}$ \citep{ritter2021sparse} & \makecell{$\mathcal{O}(NKd_{in}d_{out}+2M_{in}^3+2M_{out}^3$\\$+K(d_{out}M_{out}M_{in}+M_{in}d_{out}d_{in}))$} & $\mathcal{O}(d_{in}M_{in}+d_{out}M_{out}+2M_{in}M_{out})$ \\
\ours (Reparam) & \makecell{$\mathcal{O}(NK(d_{in}d_{out}+M_{in})+2M_{in}^3+2M_{out}^3)$}  & $\mathcal{O}(d_{in}M_{in}+d_{out}M_{out}+M_{in})$ \\
\ours (Matheron) & \makecell{$\mathcal{O}(NK(d_{in}d_{out}+M_{in})+2M_{in}^3+2M_{out}^3$\\$+K(d_{out}M_{out}M_{in}+M_{in}d_{out}d_{in}))$} & $\mathcal{O}(d_{in}M_{in}+d_{out}M_{out}+KM_{in}M_{out} + M_{in})$ \\
\bottomrule
\bottomrule
\end{tabular}
\caption{Computational complexity per layer. We assume $\mathbf{W} \in \mathbb{R}^{d_{out}\times d_{in}}$, $\mathbf{U} \in \mathbb{R}^{M_{out}\times M_{in}}$, and $K$ forward passes for each of the $N$ inputs.}
\label{tab:complexity}
\end{table*}
The pseudo-code of our approach is presented in Algorithm \ref{alg:train}.
\begin{algorithm}[!t]
\caption{Variational Training with Flow-based \emph{Posterior} over Inducing Variables}
\label{alg:train}
\begin{algorithmic}[1]
\Require Dataset $\mathcal{D}$, network $f_{\theta}$, inducing params $(m,S)$, spectral-norm flow $g_{\phi}$, hyper-params $(\lambda,\alpha,\dots)$
\Ensure Learned $\theta,\phi,m,S$
\While{not converged}
    \State \textbf{Sample base inducing} \Comment{base variational $q_{0}$}
    \State\hspace{1em} $\mathbf{u}_{0} \sim \mathcal{N}(m,S)$
    \State \textbf{Normalising-flow transform (posterior)}
    \State\hspace{1em} $(\mathbf{u},\log|{\det}J_{g}|) \gets g_{\phi}(\mathbf{u}_{0})$ \Comment{change of variables}
    \If{whitened\_u}
        \State $\mathbf{u}\gets L_{\text{row}}\mathbf{u}L_{\text{col}}^{\top}$ \Comment{jittered-Cholesky unwhitening}
    \EndIf
    \State \textbf{Kronecker weight draw}
    \State\hspace{1em} $M_{w} \gets \texttt{cond\_mean}(\mathbf{u})$ \Comment{prior-conditional maps $T_{\mathrm{row}},T_{\mathrm{col}}$}
    \State\hspace{1em} $W \gets \texttt{cg}(M_{w};\lambda)$ \Comment{conditional Gaussian (Matheron), Eq.~(\ref{eq:condition})}
    \State \textbf{Likelihood}
    \State\hspace{1em} $\mathcal{L}_{\text{loglik}} \gets \mathbb{E}\!\left[\log p(\mathcal{D}\mid f_{\theta}(\cdot;W,b))\right]$
    \State \textbf{KL—flow part} \Comment{$\mathrm{KL}(q(u)\,\|\,p(u))$}
    \State\hspace{1em} $\mathcal{L}_{\text{KL-flow}}\!\gets\!\log q_{0}(\mathbf{u}_{0})-\log|{\det}J_{g}|-\log p(\mathbf{u})$
    \State \textbf{KL—conditional part}
    \State\hspace{1em} $\mathcal{L}_{\text{KL-cond}}\!\gets\!\tfrac{D}{2}\bigl(\lambda^{2}-1-2\log\lambda\bigr)$
    \State \textbf{ELBO \& update}
    \State\hspace{1em} $\text{ELBO} \gets \mathcal{L}_{\text{loglik}}-\mathcal{L}_{\text{KL-flow}}-\mathcal{L}_{\text{KL-cond}}$
    \State\hspace{1em} $\nabla(-\text{ELBO}) \rightarrow {\tt Adam}\bigl(\theta,\phi,m,S,\lambda\bigr)$
\EndWhile
\end{algorithmic}
\end{algorithm}

\subsection{Efficient Single‐Pass ID/OoD}\label{sec:alignment}
The flow‑transformed inducing variables $\mathbf{u}$ align with In‑distribution features and diverge from Out‑of‑Distribution ones, which provides training‑free, projection‑based ID/OoD separation.

For each layer $\ell$, compute the feature–gradient vector, where in this case the gradient is taken with respect to the pseudo-label:
\begin{equation}\label{eq:z}
z_i^{(\ell)} = \mathrm{vec}\bigl(h^{(\ell)}(x_i)\bigr)\;\odot\;\mathrm{vec}\bigl(\nabla_{h^{(\ell)}}\ell(y_i,\hat y_i)\bigr)\in\mathbb{R}^{N_\ell}
\end{equation}
Project $z_i^{(\ell)}$ onto the row‐space of $U^{(\ell)}\in\mathbb{R}^{M_\ell\times N_\ell}$ by solving a regularised least‐squares problem:
\begin{equation}\label{eq:projection}
\begin{split}
\widetilde z_i^{(\ell)}
&= P^{(\ell)}\,z_i^{(\ell)},
\\
P^{(\ell)} &= \arg\min_P \|P\,U^{(\ell)} - I\|_F^2 + \lambda\|P\|_F^2
\end{split}
\end{equation}

Finally, define the OoD score as the average projection residual:
\begin{equation}\label{eq:S_def}
S = \inf_{\|x\|=1}\|(I - P)\,g\bigl(T_{\mathrm{row}}\,U\,T_{\mathrm{col}}^\top\,x\bigr)\|,
\end{equation}
\begin{mybar}
\begin{Lemma}[Spectral Residual Separation]
Let 
\begin{equation}\label{eq:W_decomp}
W = T_{\mathrm{row}}\,U\,T_{\mathrm{col}}^\top + E.
\end{equation}
Define the projector 
\begin{equation}\label{eq:P_def}
P = U^\top\bigl(UU^\top)^{-1}U,
\end{equation}
and let \(g\) be a 1‑Lipschitz flow. Set
\begin{equation}\label{eq:S_def2}
S = \inf_{\|x\|=1}\|(I - P)\,g\bigl(T_{\mathrm{row}}\,U\,T_{\mathrm{col}}^\top\,x\bigr)\|,
\end{equation}
\begin{equation}\label{eq:d_def}
d(x) = \|(I - P)\,g\bigl((T_{\mathrm{row}}\,U\,T_{\mathrm{col}}^\top + E)\,x\bigr)\|.
\end{equation}
If during training
\begin{equation}\label{eq:S_condition}
S > 2\,\|E\|,
\end{equation}
then
\begin{equation}\label{eq:sep_condition}
\sup_{x_{\rm ID}}d(x_{\rm ID}) \;<\;\inf_{x_{\rm OoD}}d(x_{\rm OoD}),
\end{equation}
strictly separating ID and OoD samples.
\end{Lemma}
\end{mybar}

\begin{algorithm}[!t]
\caption{One-Pass ID/OoD Scoring }
\label{alg:ood}
\begin{algorithmic}[1]
\Require Trained model $f_{\theta}$, key layer set $\mathcal{L}$, test batch $\{\mathbf{x}_{i}\}$, optional projector $\mathcal{P}$
\Ensure OoD scores $\{s_{i}\}$
\Statex \textbf{Pre-compute:} for each $\ell\!\in\!\mathcal{L}$ sample $\mathbf{U}^{(\ell)}$; build $P^{(\ell)}$ with Eq.~\ref{eq:projection}.
\ForAll{$\mathbf{x}_{i}$}
    \State Run one forward–backward pass to get $\mathbf{f}_{i}^{(\ell)},\mathbf{g}_{i}^{(\ell)}$
    \ForAll{$\ell\in\mathcal{L}$}
        \State $\mathbf{z}_{i}^{(\ell)}\!\gets\!(\mathbf{f}_{i}^{(\ell)}\odot\mathbf{g}_{i}^{(\ell)})$
        \If{$\mathcal{P}$ exists}\State $\mathbf{z}_{i}^{(\ell)}\!\gets\!\mathcal{P}^{(\ell)}(\mathbf{z}_{i}^{(\ell)})$\EndIf
        \State $\mathbf{r}_{i}^{(\ell)}\!\gets\!(I-P^{(\ell)})\mathbf{z}_{i}^{(\ell)}$ 
    \EndFor
    \State $s_{i}\!\gets\!\dfrac{1}{|\mathcal{L}|}\sum_{\ell}\|\mathbf{r}_{i}^{(\ell)}\|_{2}$ \Comment{Eq. (\ref{eq:S_def})}
\EndFor
\end{algorithmic}
\end{algorithm}
(The detailed proof is provided in Appendix D.)

\noindent\textbf{Hyperparameter Configuration:}
To satisfy Eq.~(\ref{eq:S_condition}), we constrain hyperparameters: We set $\lambda \le 10^{-2}$ with L2 penalty $\beta\lambda^2$ ($\|E\| \approx 0.02\!-\!0.04$); set \texttt{whitened\_u=True}, wrap all linear layers with \texttt{spectral\_norm}, and initialize $\Sigma_{NM}$ orthogonally (approximating $S \approx 0.15$). Thus, empirically,
\begin{equation}
S \approx 0.15 > 2 \times 0.03 = 0.06 \;\Longrightarrow\; S > 2\|E\|
\end{equation}
Since the OoD distribution is unknown during training, we cannot rigorously prove the bound $S > 2\|E\|$. However, based on our empirical observations, the value of $\|E\|$ consistently falls in the above-mentioned range, and the measured value of $S_{\text{OoD}}$ significantly exceeds $2\|E\|$.

%% file: section/5-experiments.tex
\section{Experiments}
We conducted comprehensive experiments on synthetic 1-D regression; image classification on ImageNet-1k and CIFAR-100, including Out-of-Distribution detection; and semantic segmentation on CamVID and CityScapes, likewise evaluating Out-of-Distribution detection. We compare our results against several state-of-the-art and relevant baselines.

\paragraph{Datasets.}
We conduct regression experiments on the Synthetic 1-D dataset, classification experiments on image datasets including CIFAR-100 and ImageNet, and semantic segmentation experiments on CamVID and CityScapes. For Out-of-Distribution (OoD) detection, we follow standard evaluation protocols using the following dataset pairs: CIFAR-100 vs. CIFAR-10, CIFAR-10 vs. ImageNet and SVHN, as well as CityScapes vs. CamVID and CamVID vs. CityScapes.
\paragraph{Model Complexity.} 

We analyze per-layer computational and storage complexity under standard assumptions: weight matrix $\mathbf{W} \in \mathbb{R}^{d_{\mathrm{out}} \times d_{\mathrm{in}}}$, inducing matrix $\mathbf{U} \in \mathbb{R}^{M_{\mathrm{out}} \times M_{\mathrm{in}}}$, $N$ inputs, and $K$ forward passes (Table~\ref{tab:complexity}). The deterministic baseline shows minimal complexity, while BatchEnsemble scales linearly with $K$. Compared to FFG-$\mathbf{U}$ \citep{ritter2021sparse}, our reparameterisation method reduces time complexity by eliminating $K$-scaled matrix products. Our Matheron approach maintains similar time complexity to FFG-$\mathbf{U}$ but requires additional negligible storage for $K$ low-rank components. Full complexity expressions are provided in Table~\ref{tab:complexity}.

\subsection{Synthetic 1-D Regression}
In synthetic 1-D regression task, we follow \citep{miyato2018spectral} and \citep{ritter2021sparse}, who took 2 input clusters \(x_1 \sim \mathcal{U}[0.5, 0.8]\), \(x_2 \sim \mathcal{U}[1.2, 1.6]\), and targets \(y \sim \mathcal{N}(\cos(4x + 0.8), 0.01)\).
The deterministic backbone is a fully connected network with 3 hidden layers of width 100. Each hidden layer consists of a linear transformation, batch normalisation, and a tanh activation to stabilise bounded outputs. 

\begin{figure}[t]
\centering
\includegraphics[width=0.45\textwidth]{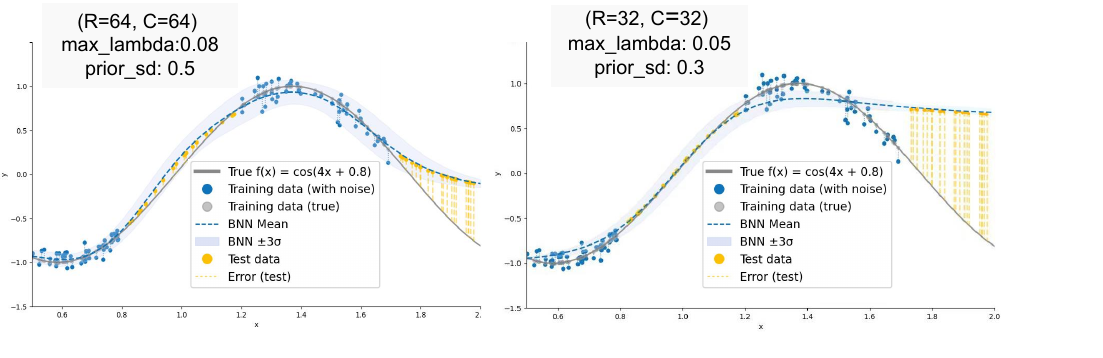}
\caption{Synthetic 1-D regression: true $f(x)=\cos(4x+0.8)$ (black); noisy training (blue) with error bars; test (orange) with error lines; \ours mean (dashed blue) and $\pm3\sigma$ interval (shaded). \textbf{Left:} inducing grid $R\times C=64\times64$ (rows$\times$columns), $\lambda_{max}=0.08$, $\mathrm{prior\_sd}=0.5$. \textbf{Right:} $R\times C=32\times32$, $\lambda_{max}=0.05$, $\mathrm{prior\_sd}=0.3$.}

\label{fig:regression}
\end{figure}

In Figure \ref{fig:regression}, we compare two different configurations that differ in the shape of their inducing matrix and the associated hyperparameters. The larger inducing matrix is equiped with higher tolerance ($\lambda$ and $\sigma$) (left side of the figure) and, therefore, outperforms the alternative (right side of the figure) as it has greater posterior expressiveness.
\subsection{Classification}
We evaluate our proposed method on standard image classification benchmarks: CIFAR-100 and ImageNet-1k. Our goal is to assess not only predictive accuracy, but also the quality of uncertainty estimation under in-distribution (ID) and Out-of-Distribution (OoD) conditions.
\par
\begin{figure}[t]
\centering
\includegraphics[width=0.3\textwidth]{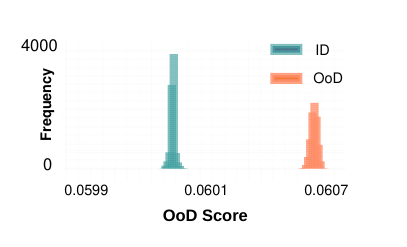}
\caption{Distributions of predictive scores from ResNet-18 + \ours (Reparam, 4 layers) on CIFAR-100 (ID) and CIFAR-10 (OoD).}
\label{fig:idood_new}
\end{figure}

\noindent\textbf{Results.}  On ImageNet-1k, Matheron sampling approach applied to all layers achieves state-of-the-art accuracy (70.19\%) while significantly reducing parameters to 5.62M (51.6\% compression versus deterministic baseline). For uncertainty estimation, all our variants demonstrate exceptional OoD detection performance, see Figure~\ref{fig:idood_new}, with Matheron sampling achieving near-perfect AUROC scores of 99.9\% on both SVHN and CIFAR-10 OoD benchmarks, substantially outperforming BatchEnsemble (94.3\%/89.9\%) and FFG-U (83.7\%/76.2\%). On CIFAR-100, the 4-layer Matheron implementation establishes new benchmarks across multiple metrics: highest accuracy (76.27\%), and most efficient parameter usage (5.51M).
\par
\begin{table*}[!ht]
\centering
\renewcommand{\arraystretch}{1.8}
\resizebox{\textwidth}{!}{
\begin{tabular}{@{}ccccccccc@{}}
\toprule
\toprule
\multicolumn{2}{c}{\textbf{Method}} & 
\textbf{Accuracy(\%) $\uparrow$} & 
\textbf{NLL} $\downarrow$ & 
\textbf{ECE (\%)} $\downarrow$ & 
\textbf{AUROC SVHN (\%)($\uparrow$)} & \textbf{AUROC CIFAR-10(\%) ($\uparrow$)} & \textbf{FLOPs}&\textbf{\parbox{4cm}{\centering \#Parameters(M) $\downarrow$}}
\\
\midrule
\multicolumn{2}{c}{Deterministic\textsuperscript{+}} & 69.68 &  1.12 & 5.25 & - & - &  \textbf{3.62G} & 11.6M \\
\multicolumn{2}{c}{BatchEnsemble\textsuperscript{+}} & 70.01 &  1.06 & \textbf{3.91}& 94.3 & 89.9 & 7.81G & 12.4M \\
\midrule
\multicolumn{2}{c}{FFG-U \citep{ritter2021sparse}	\textsuperscript{+}} & 68.31 &  0.98 & 4.17& 83.7 & 76.2 & 11.4G & 6.15M \\
\multicolumn{2}{c}{F-SGVB-LRT \citep{nguyen2024flat}} & 68.42  & 2.71 & 5.91 &-&-&-& 13.1M  \\

\midrule
\multirow{3}{*}{\ours}
  & \multicolumn{1}{c}{(Reparam, 4 layers)}     & 70.00  & 1.32 & 5.91 & 99.8 & 98.9 & 8.08G &  8.45M   \\
  & \multicolumn{1}{c}{(Matheron, 4 layers)}  & 70.17  & 1.13 & 6.86 & \textbf{99.9} & \textbf{99.9} & 8.09G  &  8.48M  \\
  & \multicolumn{1}{c}{(Matheron, all layers)}& \textbf{70.19} &1.06 & 4.81 & \textbf{99.9} & \textbf{99.9} & 14.7G  & \textbf{5.62M} \\

\bottomrule
\bottomrule
\end{tabular}
}
\caption{Comparison of \ours{} and competitive techniques on the \textbf{ImageNet-1k} dataset (superscripts indicate our reproduced variants\textsuperscript{+}, otherwise from original papers). The last three rows show our implementations using two sampling methods—\emph{Reparam} and \emph{Matheron}—applied to four convolutional layer pairs (\emph{Reparam, 2 pairs}\textsuperscript{+}, \emph{Matheron, 2 pairs}\textsuperscript{+}) and to all layers (\emph{Matheron, all layers}\textsuperscript{+}). Note: all methods are based on the \textbf{ResNet-18} architecture for fair comparison; }
\label{tab:imagenet1k}
\end{table*}



\begin{table*}[!ht]
\centering

\renewcommand{\arraystretch}{1.8}
\resizebox{\textwidth}{!}{
\begin{tabular}{@{}ccccccccc@{}}
\toprule
\toprule
\multicolumn{2}{c}{\textbf{Method}} & 
\textbf{Accuracy(\%) $\uparrow$} & 
\textbf{NLL} $\downarrow$ & 
\textbf{ECE (\%)} $\downarrow$ & \makecell{\textbf{AUROC} \\ \textbf{CIFAR-100 }$\rightarrow$ \textbf{SVHN}} & \makecell{\textbf{AUROC} \\ \textbf{CIFAR-100 }$\rightarrow$ \textbf{CIFAR-10}} & \textbf{FLOPs}&\textbf{\parbox{4cm}{\centering \#Parameters(M) $\downarrow$}}
\\
\midrule
\multicolumn{2}{c}{Deterministic\textsuperscript{+}} & 75.61 &  0.93 & 4.31& - & - &  \textbf{1.1G} & 11.2M \\
\multicolumn{2}{c}{BatchEnsemble\textsuperscript{+}} & 76.01 &  0.99 & 3.26& 91.1 & 83.4 & 4.8G & 11.9M \\
\midrule
\multicolumn{2}{c}{FFG-U \citep{ritter2021sparse}	\textsuperscript{+}} & 74.81 &  0.99 & 4.31& 80.6 & 75.4 & 11.8G & 5.85M \\
\multicolumn{2}{c}{F-SGVB-LRT \citep{nguyen2024flat}} & 70.10  & 1.12 & 3.62 &-&-&-& 11.8M  \\

\midrule
\multirow{3}{*}{\ours}
  & \multicolumn{1}{c}{(Reparam, 4 layers)}  & 75.99  & 1.15 & 4.72 & 99.8 & 98.9 & \textbf{2.4G} &  8.01M   \\
  & \multicolumn{1}{c}{(Matheron, 4 layers)}  & \textbf{76.27}  & 1.08 & 3.69 & \textbf{99.9} & \textbf{99.9} & 5.4G  & 8.01M  \\
  & \multicolumn{1}{c}{(Matheron, all layers)}& 76.11 &{0.92} & {3.60} & \textbf{99.9} & \textbf{99.9} & 11.8G  & \textbf{5.51M} \\

\bottomrule
\bottomrule
\end{tabular}
}
\caption{Comparison of \ours{} and competitive techniques on the \textbf{CIFAR-100} dataset (superscripts indicate our reproduced variants\textsuperscript{+}, otherwise from original papers). Note: all methods are based on the \textbf{ResNet-18} architecture for fair comparison.}
\label{tab:cifar100}
\end{table*}
\begin{table*}[!htp]
\centering
\renewcommand{\arraystretch}{1.3}

 \begin{adjustbox}{max width=2.1\columnwidth,center}
\begin{tabular}{lccccccc}
\toprule
\toprule
\textbf{Method} & \textbf{mIoU(\%) $\uparrow$} &\textbf{ NLL$ \downarrow$} &\textbf{MPA (\%) $\uparrow$}& \textbf{ECE(\%) $\downarrow$} & \makecell{AUROC \\ CamVID $\rightarrow$ \textbf{CityScapes}} & \textbf{FLOPs} & \textbf{\#Parameters(M)}\\
\midrule
Deterministic\textsuperscript{+} & 62.2 &0.57 &77.3 & 8.6& - & 115.6G & 35.3M  \\
BatchEnsemble\textsuperscript{+} & \textbf{63.1} &0.36&80.1& \textbf{5.2}& 84.4 & 126.2G & 42.2M \\
FFG-U \textsuperscript{+} & 60.1&0.37&77.1& 8.3 & 80.9 & 149.1G& 15.8M \\
\ours (Reparam, 4 layers) & 62.4&0.40&78.2& 7.5& \textbf{99.9} & \textbf{116.8G}& 32.7M \\
\ours (Matheron, 4 layers) & 62.8&0.31&78.9& 7.4& \textbf{99.9} & 119.5G & 32.7M \\
\ours (Matheron, all layers) & 62.6&0.48&79.2& {7.1}& \textbf{99.9} & 149.2G& 16.2M\\
\bottomrule
\bottomrule
\end{tabular}
  \end{adjustbox}
\caption{Comparison of \ours and competitive techniques on \textbf{CamVID}. Note: all methods are based on the \textbf{FCN-ResNet50} architecture for fair comparison.}
\label{tab:camvid}
\end{table*}

\begin{table*}[!htp]
\centering
\renewcommand{\arraystretch}{1.3}
\begin{adjustbox}{max width=2.1\columnwidth,center}
\begin{tabular}{lccccccc}
\toprule
\toprule
\textbf{Method} & \textbf{mIoU(\%) $\uparrow$} &\textbf{ NLL$ \downarrow$} &\textbf{MPA (\%) $\uparrow$}& \textbf{ECE(\%) $\downarrow$} & \makecell{AUROC \\ CityScapes $\rightarrow$ \textbf{CamVID}} & \textbf{FLOPs} & \textbf{\#Parameters(M)}\\
\midrule
Deterministic\textsuperscript{+} & 81.0 & 0.18 & 96.4 & 1.9& - & 373.9G & 65.8M  \\
BatchEnsemble\textsuperscript{+} & \textbf{81.5} &0.11& 97.1& 1.2& 88.6 & 394.8G & 70.2M \\
FFG-U \textsuperscript{+} & 78.1 &0.21&91.7& 3.2 & 71.1 & 459.7G& 55.7M \\
\ours (Reparam, 4 layers) & 80.4 & 0.15&96.1& 2.6& \textbf{99.9} & 374.3G& 65.9M \\
\ours (Matheron, 4 layers) & 80.9 & 0.11&96.5& 1.2& \textbf{99.9} & 374.9G & 65.9M \\
\ours (Matheron, all layers) & 80.7 & 0.18&95.2& 1.8& \textbf{99.9} & 460.3G& 58.0M\\
\bottomrule
\bottomrule
\end{tabular}
\end{adjustbox}
\caption{Evaluation of \ours against other state-of-the-art methods on \textbf{CityScapes}, with all approaches employing the \textbf{HRNet-W48} backbone for a fair comparison; Note: we applied a pre-trained model here to reduce the computational cost.}
\label{tab:CityScapes}
\end{table*}

\subsection{Semantic Segmentation}
\begin{figure}[!ht]
\centering

\includegraphics[width=0.4\textwidth]{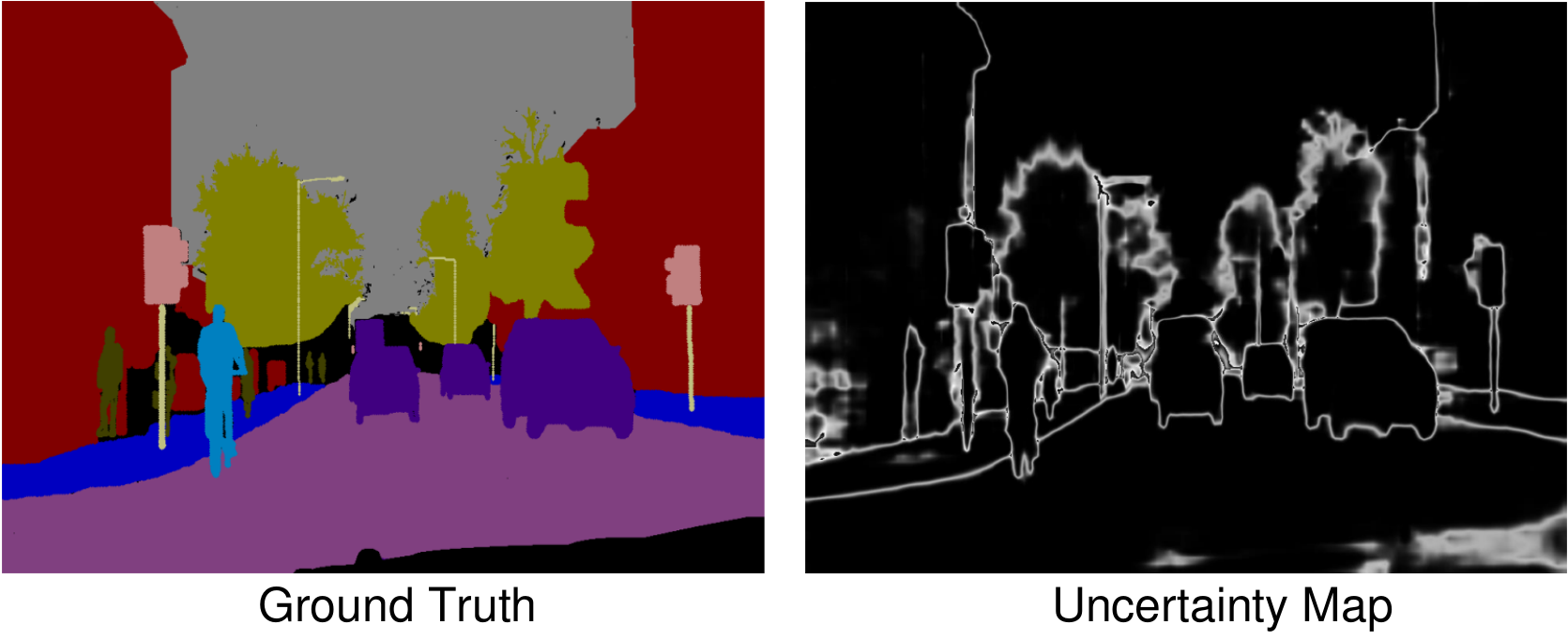}
\caption{CamVID Ground Truth (left) and Uncertainty Map (right) generated by FCN-ResNet50 (Matheron, 4 layers). }
\label{fig:semantic}
\end{figure}

\begin{figure}[!ht]
\centering

\includegraphics[width=0.4\textwidth]{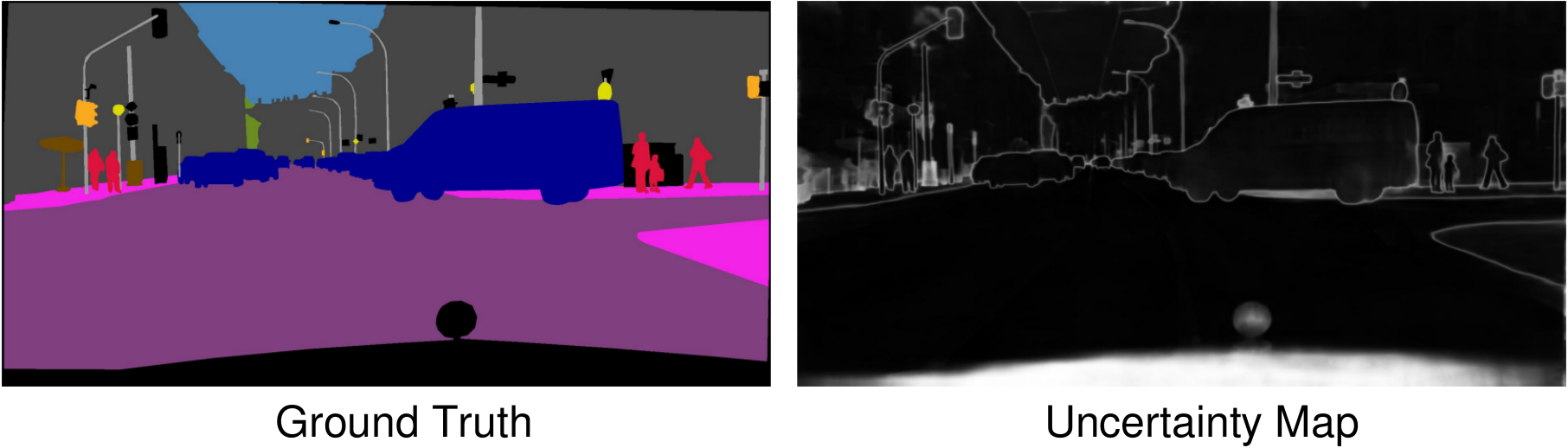}
\caption{\textbf{CityScapes} Ground Truth (left) and Uncertainty Map (right) generated by HRNet-W48 (Matheron, 4 layers). }
\label{fig:CityScapes}
\end{figure}

For the CamVID semantic segmentation task, we evaluated \ours using FCN-ResNet50 as the backbone. Our method delivered competitive results (Table \ref{tab:camvid}): the 4 layers Matheron variant achieved an mIoU of 62.9\%, which is very close to BatchEnsemble's state-of-the-art performance of 63.1\%. Moreover, all configurations of \ours exhibited remarkable robustness against domain shifts, attaining a 99.9\% AUROC for the CamVID→CityScapes generalization task.
We also conducted experiments on CityScapes, using the HRNet-W48 as backbone. Table \ref{tab:CityScapes} shows that our method demonstrates superior generalisation ability. It achieved a high mIoU of 80.9\% and an AUROC of 99.9\% for the CityScapes to CamVID domain shift task, along with competitive NLL (0.11) and ECE (1.2\%).
In both experiments, we not only presented the predicted segmentation results but also visualised the associated uncertainty distributions. These distributions clearly indicate increased uncertainty along object boundaries, for CamVID (Figure \ref{fig:semantic}) and CityScapes (Figure \ref{fig:CityScapes}).
\subsection{Ablation Experiments}
The shape of the inducing matrix in \ours determines a trade-off between model compression and accuracy. To quantify this effect, we conduct ablation studies varying the inducing matrix size on CIFAR-100 using ResNet-18. As shown in Table~\ref{tab:ablation}, larger matrices generally achieve higher accuracy at the cost of fewer parameter savings. The $256 \times 256$ configuration ($256\times256$) achieves the highest accuracy (77.48\%) without compression, while the smallest ($32\times32$) achieves maximum compression (87.9\%) with reduced accuracy (69.83\%). The $128\times128$ setting provides a favourable balance, which we adopted throughout our experiments. We always set the linear layer to 128 × num\_class.

\begin{table}[!h]
\centering
\renewcommand{\arraystretch}{1.3}
\caption{CIFAR‑100 results for ResNet‑18 modified with \ours: accuracy, parameter count, and compression rate relative to the 11.2 M‑parameter deterministic model, over various inducing‑matrix sizes.}
\begin{adjustbox}{max width=0.97\columnwidth,center}
\begin{tabular}{lccccc}
\toprule
\multicolumn{2}{c}{Type/Size} & \textbf{Accuracy(\%) $\uparrow$} & 
\textbf{NLL} $\downarrow$ & 
\textbf{ECE (\%)} $\downarrow$ &  \makecell{Parameters \\Compression}$\uparrow$\\
\midrule
\midrule
\multirow{4}{*}{Conv}& $32\times32$ &69.83&1.16& 4.46& 1.35M / 87.9\% \\
&$64\times64$&73.26&1.14& 3.88 & 2.66M / 76.2\% \\
&$\mathbf{128\times128}$&\textbf{76.11}&\textbf{0.92}& \textbf{3.60}& \textbf{5.51M / 50.8\%}  \\
&$256\times256$&77.48&1.00& 4.55& 12.1M / -\phantom{0.00\%} \\
\midrule
\multicolumn{2}{c}{Linear setting} & \multicolumn{4}{c}{$128 \times$ \text{num\_class}}  \\
\bottomrule
\end{tabular}
\end{adjustbox}
\label{tab:ablation}
\end{table}

%% file: section/6-discussion.tex
\section{Discussion}
\textbf{Layer Selection:} In our experiments, we selected 2 pairs (4 layers) of consecutive convolution layers for replacement with \ours. Furthermore, the second layer in each pair will be used to compute the predictive score to distinguish ID from OoD data, as the output of the first layer satisfies the 1-Lipschitz condition (see Appendix C for details). 
While these 2 pairs can be chosen randomly, selecting one layer from shallow layers and another one from deep layers can reduce the computational cost and increase the separation distance between ID and OoD. For example, in ResNet-18, we used the layers ["layer2.1.conv1", "layer2.1.conv2", "layer4.1.conv1", "layer4.1.conv2"], where the first two layers form the first pair and the last two layers form the second pair.

%% file: section/6.5-conclusion.tex
\section{Conclusion}
We have proposed a GP-inspired uncertainty approach, named \ours, which can seamlessly be integrated into a deep neural network. \ours incorporates a compact inducing weight matrix to project neural network weights' uncertainty into a lower-dimensional subspace, with expressiveness augmented through a normalizing-flow variational posterior and spectral regularization, which aligns the inducing subspace with feature-gradient geometry through a numerically stable projection mechanism objective, and produces a single-pass projection for Out-of-Distribution (OoD) detection.
\par







%% file: section/8-checklist.tex
\setlength{\leftmargini}{20pt}
\makeatletter\def\@listi{\leftmargin\leftmargini \topsep .5em \parsep .5em \itemsep .5em}
\def\@listii{\leftmargin\leftmarginii \labelwidth\leftmarginii \advance\labelwidth-\labelsep \topsep .4em \parsep .4em \itemsep .4em}
\def\@listiii{\leftmargin\leftmarginiii \labelwidth\leftmarginiii \advance\labelwidth-\labelsep \topsep .4em \parsep .4em \itemsep .4em}\makeatother

\setcounter{secnumdepth}{0}
\renewcommand\thesubsection{\arabic{subsection}}
\renewcommand\labelenumi{\thesubsection.\arabic{enumi}}

\newcounter{checksubsection}
\newcounter{checkitem}[checksubsection]

\newcommand{\checksubsection}[1]{%
  \refstepcounter{checksubsection}%
  \paragraph{\arabic{checksubsection}. #1}%
  \setcounter{checkitem}{0}%
}

\newcommand{\checkitem}{%
  \refstepcounter{checkitem}%
  \item[\arabic{checksubsection}.\arabic{checkitem}.]%
}
\newcommand{\question}[2]{\normalcolor\checkitem #1 #2 \color{blue}}
\newcommand{\ifyespoints}[1]{\makebox[0pt][l]{\hspace{-15pt}\normalcolor #1}}

\section{Reproducibility Checklist}

\vspace{1em}
\hrule
\vspace{1em}








\checksubsection{General Paper Structure}
\begin{itemize}

\question{Includes a conceptual outline and/or pseudocode description of AI methods introduced}{(yes/partial/no/NA)}
yes

\question{Clearly delineates statements that are opinions, hypothesis, and speculation from objective facts and results}{(yes/no)}
yes

\question{Provides well-marked pedagogical references for less-familiar readers to gain background necessary to replicate the paper}{(yes/no)}
yes

\end{itemize}
\checksubsection{Theoretical Contributions}
\begin{itemize}

\question{Does this paper make theoretical contributions?}{(yes/no)}
yes

	\ifyespoints{\vspace{1.2em}If yes, please address the following points:}
        \begin{itemize}
	
	\question{All assumptions and restrictions are stated clearly and formally}{(yes/partial/no)}
	yes

	\question{All novel claims are stated formally (e.g., in theorem statements)}{(yes/partial/no)}
	yes

	\question{Proofs of all novel claims are included}{(yes/partial/no)}
	partial

	\question{Proof sketches or intuitions are given for complex and/or novel results}{(yes/partial/no)}
	yes

	\question{Appropriate citations to theoretical tools used are given}{(yes/partial/no)}
	yes

	\question{All theoretical claims are demonstrated empirically to hold}{(yes/partial/no/NA)}
	yes

	\question{All experimental code used to eliminate or disprove claims is included}{(yes/no/NA)}
	yes
	
	\end{itemize}
\end{itemize}

\checksubsection{Dataset Usage}
\begin{itemize}

\question{Does this paper rely on one or more datasets?}{(yes/no)}
yes

\ifyespoints{If yes, please address the following points:}
\begin{itemize}

	\question{A motivation is given for why the experiments are conducted on the selected datasets}{(yes/partial/no/NA)}
	yes

	\question{All novel datasets introduced in this paper are included in a data appendix}{(yes/partial/no/NA)}
	yes

	\question{All novel datasets introduced in this paper will be made publicly available upon publication of the paper with a license that allows free usage for research purposes}{(yes/partial/no/NA)}
	yes

	\question{All datasets drawn from the existing literature (potentially including authors' own previously published work) are accompanied by appropriate citations}{(yes/no/NA)}
	yes

	\question{All datasets drawn from the existing literature (potentially including authors' own previously published work) are publicly available}{(yes/partial/no/NA)}
	yes

	\question{All datasets that are not publicly available are described in detail, with explanation why publicly available alternatives are not scientifically satisficing}{(yes/partial/no/NA)}
	yes

\end{itemize}
\end{itemize}

\checksubsection{Computational Experiments}
\begin{itemize}

\question{Does this paper include computational experiments?}{(yes/no)}
yes

\ifyespoints{If yes, please address the following points:}
\begin{itemize}

	\question{This paper states the number and range of values tried per (hyper-) parameter during development of the paper, along with the criterion used for selecting the final parameter setting}{(yes/partial/no/NA)}
	yes

	\question{Any code required for pre-processing data is included in the appendix}{(yes/partial/no)}
	yes

	\question{All source code required for conducting and analyzing the experiments is included in a code appendix}{(yes/partial/no)}
	yes

	\question{All source code required for conducting and analyzing the experiments will be made publicly available upon publication of the paper with a license that allows free usage for research purposes}{(yes/partial/no)}
	yes
        
	\question{All source code implementing new methods have comments detailing the implementation, with references to the paper where each step comes from}{(yes/partial/no)}
	yes

	\question{If an algorithm depends on randomness, then the method used for setting seeds is described in a way sufficient to allow replication of results}{(yes/partial/no/NA)}
	yes

	\question{This paper specifies the computing infrastructure used for running experiments (hardware and software), including GPU/CPU models; amount of memory; operating system; names and versions of relevant software libraries and frameworks}{(yes/partial/no)}
	yes

	\question{This paper formally describes evaluation metrics used and explains the motivation for choosing these metrics}{(yes/partial/no)}
	yes

	\question{This paper states the number of algorithm runs used to compute each reported result}{(yes/no)}
	yes

	\question{Analysis of experiments goes beyond single-dimensional summaries of performance (e.g., average; median) to include measures of variation, confidence, or other distributional information}{(yes/no)}
	yes

	\question{The significance of any improvement or decrease in performance is judged using appropriate statistical tests (e.g., Wilcoxon signed-rank)}{(yes/partial/no)}
	no

	\question{This paper lists all final (hyper-)parameters used for each model/algorithm in the paper’s experiments}{(yes/partial/no/NA)}
	yes

\end{itemize}
\end{itemize}

%% file: section/6-appendix.tex
\appendix

\clearpage

\subsection*{Appendix A: Derivation of Conditional Gaussian via Completing the Square}
\label{appendix:conditional_gaussian_square}

This appendix presents a rigorous derivation of the conditional Gaussian distribution using the method of completing the square in the exponent. We emphasize the geometric interpretation of covariance matrices and demonstrate how the Schur complement naturally emerges during the process. This approach reveals the deep connection between joint and conditional distributions in Gaussian systems.

\subsubsection{Preliminary Setup}
Let $(\mathbf{x},\mathbf{y}) \in \mathbb{R}^{n+m}$ be jointly Gaussian random vectors with partitioned moments:
\begin{equation}
\begin{split}
\boldsymbol{\mu} &= \begin{bmatrix}
\boldsymbol{\mu}_x^{(n\times 1)} \\
\boldsymbol{\mu}_y^{(m\times 1)}
\end{bmatrix}, \\
\Sigma &= \begin{bmatrix}
\Sigma_{xx}^{(n\times n)} & \Sigma_{xy}^{(n\times m)} \\
\Sigma_{yx}^{(m\times n)} & \Sigma_{yy}^{(m\times m)}
\end{bmatrix}
\end{split}
\end{equation}
where superscripts in parentheses indicate matrix dimensions. The joint density is:
\begin{equation}
\begin{split}
p(\mathbf{x},\mathbf{y}) &= (2\pi)^{-\frac{n+m}{2}}|\Sigma|^{-\frac{1}{2}} \\
&\quad\times \exp\left(-\frac{1}{2}
\begin{bmatrix}
\mathbf{x} - \boldsymbol{\mu}_x \\
\mathbf{y} - \boldsymbol{\mu}_y
\end{bmatrix}^\top
\Sigma^{-1}
\begin{bmatrix}
\mathbf{x} - \boldsymbol{\mu}_x \\
\mathbf{y} - \boldsymbol{\mu}_y
\end{bmatrix}\right)
\end{split}
\end{equation}

\subsubsection{Inverse Covariance Structure}
The key insight comes from the block matrix inversion formula:
{\small
\begin{equation}
\begin{split}
\Sigma^{-1} &= \begin{bmatrix}
A & B \\
B^\top & D
\end{bmatrix} \\
=& \begin{bmatrix}
(\Sigma/\Sigma_{yy})^{-1} & -(\Sigma/\Sigma_{yy})^{-1}\Sigma_{xy}\Sigma_{yy}^{-1} \\
-\Sigma_{yy}^{-1}\Sigma_{yx}(\Sigma/\Sigma_{yy})^{-1} & (\Sigma_{yy} - \Sigma_{yx}\Sigma_{xx}^{-1}\Sigma_{xy})^{-1}
\end{bmatrix}
\end{split}
\end{equation}
}
where $\Sigma/\Sigma_{yy} \coloneqq \Sigma_{xx} - \Sigma_{xy}\Sigma_{yy}^{-1}\Sigma_{yx}$ denotes the Schur complement. This reveals the fundamental relationship:
\begin{equation}
A^{-1} = \Sigma_{xx} - \Sigma_{xy}\Sigma_{yy}^{-1}\Sigma_{yx}
\end{equation}

Expanding the Mahalanobis distance in the exponent:
\begin{equation}
\begin{split}
\mathcal{Q} &\coloneqq
\begin{bmatrix}
\mathbf{x} - \boldsymbol{\mu}_x \\
\mathbf{y} - \boldsymbol{\mu}_y
\end{bmatrix}^{\!\top}
\begin{bmatrix}
A & B \\
B^\top & D
\end{bmatrix}
\begin{bmatrix}
\mathbf{x} - \boldsymbol{\mu}_x \\
\mathbf{y} - \boldsymbol{\mu}_y
\end{bmatrix} \\[6pt]
&= (\mathbf{x} - \boldsymbol{\mu}_x)^\top A (\mathbf{x} - \boldsymbol{\mu}_x) \\
&\quad+ 2(\mathbf{x} - \boldsymbol{\mu}_x)^\top B (\mathbf{y} - \boldsymbol{\mu}_y) \\
&\quad+ (\mathbf{y} - \boldsymbol{\mu}_y)^\top D (\mathbf{y} - \boldsymbol{\mu}_y)
\end{split}
\end{equation}

\subsubsection{Conditional Density Derivation}
\begin{equation}
\begin{split}
p(\mathbf{x}|\mathbf{y}) &\propto \exp\left(-\frac{1}{2}\left[(\mathbf{x} - \boldsymbol{\mu}_x)^\top A (\mathbf{x} - \boldsymbol{\mu}_x)\right.\right. \\
&\quad\quad\left.\left.+ 2(\mathbf{x} - \boldsymbol{\mu}_x)^\top B (\mathbf{y} - \boldsymbol{\mu}_y)\right]\right)
\end{split}
\end{equation}

To complete the square, we seek $\mathbf{m} \in \mathbb{R}^n$ and $\Phi \in \mathbb{R}^{n\times n}$ such that:
\begin{equation}
\begin{split}
&(\mathbf{x} - \mathbf{m})^\top \Phi (\mathbf{x} - \mathbf{m}) = (\mathbf{x} - \boldsymbol{\mu}_x)^\top A (\mathbf{x} - \boldsymbol{\mu}_x) \\
&\quad\quad+ 2(\mathbf{x} - \boldsymbol{\mu}_x)^\top B (\mathbf{y} - \boldsymbol{\mu}_y)
\end{split}
\end{equation}

Matching coefficients yields:
\begin{subequations}
\begin{align}
\Phi &= A \\
\Phi(\mathbf{m} - \boldsymbol{\mu}_x) &= -B(\mathbf{y} - \boldsymbol{\mu}_y)
\end{align}
\end{subequations}
Solving this system gives the conditional parameters:
\begin{equation}
\mathbf{m} = \boldsymbol{\mu}_x - A^{-1}B(\mathbf{y} - \boldsymbol{\mu}_y), \quad \Phi = A
\end{equation}

\subsubsection{Canonical Form Conversion}
Using the matrix inversion lemma:
\begin{equation}
\begin{split}
A^{-1} &= \Sigma_{xx} - \Sigma_{xy}\,\Sigma_{yy}^{-1}\,\Sigma_{yx},\\
A^{-1}B  &= \Sigma_{xy}\,\Sigma_{yy}^{-1}\,
\end{split}
\end{equation}
Substituting these into the conditional parameters produces the standard form:
\begin{equation}
\begin{split}
p(\mathbf{x}\mid\mathbf{y})
&= \mathcal{N}\bigl(\mathbf{x}\mid
\underbrace{\boldsymbol{\mu}_x + \Sigma_{xy}\,\Sigma_{yy}^{-1}(\mathbf{y}-\boldsymbol{\mu}_y)}_{\text{Conditional Mean}},\\
&\quad\underbrace{\Sigma_{xx} - \Sigma_{xy}\,\Sigma_{yy}^{-1}\,\Sigma_{yx}}_{\text{Conditional Covariance}}\bigr)
\end{split}
\end{equation}
\begin{equation}
\boxed{
\begin{split}
p(\mathbf{x}\mid\mathbf{y})
&= \mathcal{N}\bigl(\mathbf{x};\,\boldsymbol{\mu}_{x|y},\,\Sigma_{x|y}\bigr),\\
\boldsymbol{\mu}_{x|y}
&= \boldsymbol{\mu}_x + \Sigma_{xy}\,\Sigma_{yy}^{-1}(\mathbf{y}-\boldsymbol{\mu}_y),\\
\Sigma_{x|y}
&= \Sigma_{xx} - \Sigma_{xy}\,\Sigma_{yy}^{-1}\,\Sigma_{yx}.
\end{split}
}
\end{equation}

\subsection*{Appendix B: Derivation of Conditional Mean via Kronecker Factorization under Matrix Normal Distributions}
\label{appendix:conditional_mean_kronecker}

In this section, we provide a detailed derivation of the conditional mean formula for jointly Gaussian-distributed matrix-valued random variables. Suppose we have random matrices $W \in \mathbb{R}^{m \times n}$ and $U \in \mathbb{R}^{p \times q}$, jointly Gaussian distributed with
\begin{equation}
\begin{split}
W &\sim \mathcal{MN}\bigl(\mathbf{0},\,\Sigma_{\mathrm{row}},\,\Sigma_{\mathrm{col}}\bigr), \\
U &\sim \mathcal{MN}\bigl(\mathbf{0},\,\Sigma_{U}^{(\mathrm{row})},\,\Sigma_{U}^{(\mathrm{col})}\bigr)
\end{split}
\end{equation}
together with the cross-covariance
\begin{equation}
\mathrm{Cov}(W,U) = \Sigma_{W,U}
\end{equation}

Given the joint Gaussian distribution, the conditional expectation of $W$ given $U$ is expressed as
\begin{equation}
\mathbb{E}[W \mid U] = \Sigma_{W,U}\,\Sigma_U^{-1}\,U
\end{equation}
where $\Sigma_{W,U}$ denotes the cross-covariance between the matrices $W$ and $U$, and $\Sigma_U$ is the covariance of $U$.

By vectorizing matrices in a column-wise fashion, we define $\mathrm{vec}(W) \in \mathbb{R}^{mn\times 1}$ and $\mathrm{vec}(U) \in \mathbb{R}^{pq\times 1}$. Consequently, the conditional expectation becomes
\begin{equation}
\begin{split}
\mathbb{E}[\mathrm{vec}(W)\mid U] &= K_{W,U}\,K_U^{-1}\,\mathrm{vec}(U), \\
K_{W,U}&\in\mathbb{R}^{mn\times pq},\quad K_U\in\mathbb{R}^{pq\times pq}
\end{split}
\end{equation}
This formulation clearly illustrates the fundamental linear mapping $K_{W,U}K_U^{-1}$ characteristic of conditional Gaussian distributions.

\subsubsection{Kronecker-Structured Covariance Decomposition}
Further assuming a Kronecker separable covariance structure, we have
\begin{equation}
\begin{split}
K_W &= \Sigma_{\mathrm{col}}\otimes \Sigma_{\mathrm{row}}, \\
K_U &= \Sigma_{U}^{(\mathrm{col})}\otimes\Sigma_{U}^{(\mathrm{row})},\\
K_{W,U}&= K_{W,U}^{(\mathrm{col})}\otimes K_{W,U}^{(\mathrm{row})}
\end{split}
\end{equation}

Using properties of the Kronecker product, specifically
\begin{equation}
\begin{split}
(A\otimes B)(C\otimes D) &= (AC)\otimes (BD), \\
(A\otimes B)^{-1} &= A^{-1}\otimes B^{-1}
\end{split}
\end{equation}
we obtain
\begin{equation}
\begin{split}
K_{W,U}K_U^{-1} = \bigl(K_{W,U}^{(\mathrm{col})}(\Sigma_{U}^{(\mathrm{col})})^{-1}\bigr)
\otimes \\
\bigl(K_{W,U}^{(\mathrm{row})}(\Sigma_{U}^{(\mathrm{row})})^{-1}\bigr)
\end{split}
\end{equation}

Thus, the conditional expectation in vectorized form is
\begin{equation}
\begin{split}
\mathrm{vec}\bigl(\mathbb{E}[W\mid U]\bigr)
= \bigl(K_{W,U}^{(\mathrm{col})}(\Sigma_{U}^{(\mathrm{col})})^{-1}\bigr)
\otimes \\
\bigl(K_{W,U}^{(\mathrm{row})}(\Sigma_{U}^{(\mathrm{row})})^{-1}\bigr)
\;\mathrm{vec}(U)
\end{split}
\end{equation}

We leverage the vectorization identity
\begin{equation}
\mathrm{vec}(A X B^\top) = (B\otimes A)\,\mathrm{vec}(X)
\end{equation}
To reconstruct the conditional expectation back to matrix form. Let us define the transformations
\begin{equation}
\begin{split}
T_{\mathrm{row}} &= K_{W,U}^{(\mathrm{row})}(\Sigma_{U}^{(\mathrm{row})})^{-1},\\
T_{\mathrm{col}} &= K_{W,U}^{(\mathrm{col})}(\Sigma_{U}^{(\mathrm{col})})^{-1}
\end{split}
\end{equation}

Then, the conditional expectation of $W$ given $U$ can be succinctly written in matrix form as
\begin{equation}
\mathbb{E}[W\mid U] = T_{\mathrm{row}}\,U\,T_{\mathrm{col}}^\top
\end{equation}

To avoid direct computation of inverses, numerical implementations typically utilize the Cholesky decomposition of covariance matrices. Specifically, for covariance matrices:
\begin{equation}
\begin{split}
\Sigma_{U}^{(\mathrm{row})} &= L_{\mathrm{row}}L_{\mathrm{row}}^\top,\\
\Sigma_{U}^{(\mathrm{col})} &= L_{\mathrm{col}}L_{\mathrm{col}}^\top
\end{split}
\end{equation}
the transformations can be computed efficiently by solving linear systems involving these Cholesky factors, thus improving numerical stability and computational efficiency.

\subsection*{Appendix C: Derivation of the Flow-based KL Divergence Term}
\newcommand{\ud}{\mathrm{d}}

Let \(u_0 \in \mathbb{R}^{RC}\) be a \emph{base} latent with density \(q_0(u_0)\), and let \(g:\mathbb{R}^{RC}\to\mathbb{R}^{RC}\) be a bijective, differentiable transformation with Jacobian \(J_g(u_0)=\partial g(u_0)/\partial u_0\). Define \(u = g(u_0)\) and the \emph{pushforward} density \(q(u)=g_{\#}q_0\). A prior density \(p(u)\) is specified on the same space as \(u\). We wish to compute
{\small
\begin{equation}
\begin{aligned}
\mathrm{KL}\!\left[q(u)\,\|\,p(u)\right]
&= \int q(u)\,
    \Bigl[\log q(u) - \log p(u)\Bigr]\,
    \ud u
\end{aligned}
\end{equation}
}
Because \(g\) is bijective we may write \(u_0 = g^{-1}(u)\) and apply the change-of-variables formula:
\begin{equation}
\begin{aligned}
q(u)
&= q_0\!\bigl(g^{-1}(u)\bigr)\;
   \Bigl|\det J_{g^{-1}}(u)\Bigr| \\
&= q_0(u_0)\;
   \Bigl|\det J_g(u_0)\Bigr|^{-1}
\end{aligned}
\end{equation}
Taking logs gives
\begin{equation}
\log q(u)
= \log q_0(u_0) - \log\!\bigl|\det J_g(u_0)\bigr|
\end{equation}

Substitute this expression for \(\log q(u)\) into the KL definition and simultaneously change integration variables from \(u\) to \(u_0\) (with \(\ud u = |\det J_g(u_0)|\,\ud u_0\), which cancels the reciprocal Jacobian factor already present in \(q(u)\)):
\begin{equation}
\begin{aligned}
\mathrm{KL}\!\left[q(u)\,\|\,p(u)\right]
= \int q_0(u_0)\,
\Bigl[
  \log q_0(u_0)
  - \\  \log\!\bigl|\det J_g(u_0)\bigr| 
  - \log p\!\bigl(g(u_0)\bigr)
\Bigr]\,
\ud u_0
\end{aligned}
\end{equation}

Recognizing the integral as an expectation over \(u_0 \sim q_0\) yields the desired identity:
\begin{equation}
\begin{aligned}
\mathrm{KL}\!\left[q(u)\,\|\,p(u)\right]
= \mathbb{E}_{u_0 \sim q_0}
  \Bigl[
    \log q_0(u_0)
    -  \\ \log\!\bigl|\det J_g(u_0)\bigr| 
    - \log p\!\bigl(g(u_0)\bigr)
  \Bigr]
\end{aligned}
\label{eq:kl_flow}
\end{equation}

\textbf{Monte Carlo estimator.} Drawing a reparameterized sample \(u_0 \sim q_0\) and computing \(u=g(u_0)\) and \(\log|\det J_g(u_0)|\) from the flow, a single-sample stochastic estimator of the KL term in \eqref{eq:kl_flow} is
\begin{equation}
\begin{aligned}
\widehat{\mathrm{KL}}
&= \log q_0(u_0)
  - \log\!\bigl|\det J_g(u_0)\bigr| \\
&\qquad
  - \log p\!\bigl(g(u_0)\bigr)
\end{aligned}
\end{equation}
optionally averaged over multiple samples to reduce variance.

\textbf{Adding a conditional Gaussian factor (weights).} Suppose model weights \(W \in \mathbb{R}^{D}\) (flattened) are conditionally Gaussian given \(u\):
\begin{equation}
W \mid u \sim \mathcal{N}\!\bigl(\mu(u), (\lambda\sigma_p)^2 I\bigr),
\end{equation}
with Gaussian prior
\begin{equation}
W \sim \mathcal{N}\!\bigl(\mu(u), \sigma_p^2 I\bigr),
\end{equation}
---note the same mean (the prior mean may be zero in practice; if means differ, include the usual quadratic term).
Then the per-\(u\) conditional KL is the sum over \(D\) independent dimensions of the 1D Gaussian KL:
{\small
\begin{equation}
\begin{aligned}
\mathrm{KL}\!\Bigl[
  \mathcal{N}\bigl(\mu, (\lambda\sigma_p)^2\bigr)
  \,\Big\|\,
  \mathcal{N}\bigl(\mu, \sigma_p^2\bigr)
\Bigr]
= \frac{1}{2}\Bigl(\lambda^2 - 1 - 2\log\lambda\Bigr)
\end{aligned}
\end{equation}
hence
\begin{equation}
\mathrm{KL}\!\left[q(W\mid u)\,\|\,p(W\mid u)\right]
= \frac{D}{2}\Bigl(\lambda^2 - 1 - 2\log\lambda\Bigr)
\label{eq:kl_conditional_weights}
\end{equation}
}
Combining \eqref{eq:kl_flow} and \eqref{eq:kl_conditional_weights} yields the total variational KL used in the ELBO of the inducing-parameter model described in the main text.

\subsection*{Appendix D: Proof of Spectral Residual Separation}

\textbf{Lemma (Spectral Residual Separation)}  
Let the weight matrix be decomposed as
\begin{equation}
w = T_{\mathrm{row}} U T_{\mathrm{col}}^\top + E
\end{equation}
Define the separation margin and residual respectively as
\begin{equation}
S = \inf_{\|x\|=1}\|(I - P)\, g(T_{\mathrm{row}} U T_{\mathrm{col}}^\top x)\|
\end{equation}
\begin{equation}
d(x) = \|(I - P)\, g((T_{\mathrm{row}} U T_{\mathrm{col}}^\top + E)x)\|
\end{equation}
where the projection $P$ is given by
\begin{equation}
P = U^\top (U U^\top + \lambda I)^{-1} U
\end{equation}
with $\|I - P\|_\sigma = 1$, and $g:\mathbb{R}^n\rightarrow\mathbb{R}^n$ is 1–Lipschitz.

For in-distribution (ID) samples, we have empirically:
\begin{equation}
\sup_{x_{\mathrm{ID}}} d(x_{\mathrm{ID}}) < \|E\|
\end{equation}

If during training, we achieve (empirically observed) for Out-of-Distribution (OoD) samples:
\begin{equation}
S_{\mathrm{OoD}} > 2\,\|E\|,
\end{equation}
then the following strict separation holds:
\begin{equation}
\sup_{x_{\mathrm{ID}}} d(x_{\mathrm{ID}}) < \inf_{x_{\mathrm{OoD}}} d(x_{\mathrm{OoD}})
\end{equation}

\medskip
\noindent\textbf{Proof.}  
\medskip
We divided the proof into 4 steps: 

\noindent\textbf{Step 1: Spectral norm of $I - P$ when $U$ is square but rank-deficient (ignoring $\lambda I$)}\\[4pt]
Suppose \(U \in \mathbb{R}^{N \times N}\) is a square matrix with rank \(r < N\), so it is not invertible.  
We consider the projection matrix defined by
\begin{equation}
P = U^{\top}(UU^{\top})^{\dagger}U,
\end{equation}
where \((\cdot)^{\dagger}\) denotes the Moore–Penrose pseudoinverse.

Using the singular value decomposition \(U = V \Sigma W^{\top}\), with
\begin{equation}
\Sigma = \mathrm{diag}(\sigma_1, \dots, \sigma_r, 0, \dots, 0),
\quad \text{where } \sigma_1 \ge \dots \ge \sigma_r > 0
\end{equation}
we can express \(P\) as
\begin{equation}
P = W \,
\mathrm{diag}(\underbrace{1, \dots, 1}_{r}, \underbrace{0, \dots, 0}_{N-r}) \,
W^{\top}
\end{equation}
Therefore,
\begin{equation}
I - P = W \,
\mathrm{diag}(\underbrace{0, \dots, 0}_{r}, \underbrace{1, \dots, 1}_{N-r}) \,
W^{\top}
\end{equation}

This shows that \(I - P\) has eigenvalue 1 in the null space of \(U\) (which is of dimension \(N - r > 0\)) and eigenvalue 0 in the row space of \(U\).  
Since \(I - P\) is symmetric, its spectral norm equals its largest eigenvalue magnitude:
\begin{equation}
\|I - P\|_\sigma = \max_{\lambda \in \mathrm{spec}(I - P)} |\lambda| = 1.
\end{equation}

\textit{Remark:} If \(U\) were full-rank and invertible, then
\begin{equation}
UU^{\top} = UU^{\top} = U U^{\top} \Rightarrow (UU^{\top})^{-1} = U^{-\top} U^{-1}
\end{equation}
so
\begin{equation}
P = U^{\top} (UU^{\top})^{-1} U = U^{\top} U^{-\top} U^{-1} U = I
\end{equation}
and thus
\begin{equation}
\|I - P\|_\sigma = \|0\|_\sigma = 0
\end{equation}
Therefore, \(\|I - P\|_\sigma = 1\) if and only if \(U\) is not full rank.
\medskip

\noindent\textbf{Step 2: Upper bound for ID residuals:}  
For any in-distribution sample $x_{\mathrm{ID}}$, by construction:
\begin{equation}
(I-P)\,g(T_{\mathrm{row}} U T_{\mathrm{col}}^\top x_{\mathrm{ID}}) = 0
\end{equation}
Thus,
\begin{equation}
\begin{aligned}
d(x_{\mathrm{ID}}) 
&= \|(I - P)\,[g((T_{\mathrm{row}} U T_{\mathrm{col}}^\top + E)x_{\mathrm{ID}}) \\
&\qquad\quad - g(T_{\mathrm{row}} U T_{\mathrm{col}}^\top x_{\mathrm{ID}})]\| \\
&\le \|I - P\|_{\sigma}\,\|g((T_{\mathrm{row}} U T_{\mathrm{col}}^\top + E)x_{\mathrm{ID}}) \\
&\qquad\quad - g(T_{\mathrm{row}} U T_{\mathrm{col}}^\top x_{\mathrm{ID}})\| \\
&\le \|E\, x_{\mathrm{ID}}\|\quad(\text{$g$ is 1–Lipschitz})\\
&\le \|E\|
\end{aligned}
\end{equation}
Therefore,
\begin{equation}
\sup_{x_{\mathrm{ID}}} d(x_{\mathrm{ID}}) \le \|E\|
\end{equation}
\textbf{Explanation of critical steps:} 
The penultimate inequality holds because $g$ being 1-Lipschitz implies:
\begin{equation}
\|g(\mathbf{a}) - g(\mathbf{b})\| \leq \|\mathbf{a} - \mathbf{b}\| \quad \forall \mathbf{a},\mathbf{b}
\end{equation}
Here we set $\mathbf{a} = (T_{\mathrm{row}} U T_{\mathrm{col}}^\top + E)x_{\mathrm{ID}}$ and $\mathbf{b} = T_{\mathrm{row}} U T_{\mathrm{col}}^\top x_{\mathrm{ID}}$, yielding:
\begin{equation}
\|\mathbf{a} - \mathbf{b}\| = \|E x_{\mathrm{ID}}\|
\end{equation}
The final step uses the Cauchy-Schwarz inequality $\|E x_{\mathrm{ID}}\| \leq \|E\|\cdot\|x_{\mathrm{ID}}\|$ combined with the unit norm constraint $\|x_{\mathrm{ID}}\|=1$ from the infimum definition in Lemma 1.

Therefore,
\begin{equation}
\sup_{x_{\mathrm{ID}}} d(x_{\mathrm{ID}}) \le \|E\|
\end{equation}
\medskip
\noindent\textbf{Step 3: Lower bound for OoD residuals:}  
For any Out-of-Distribution vector $x_{\mathrm{OoD}}$, by reverse triangle inequality,
\begin{equation}
\begin{aligned}
d(x_{\mathrm{OoD}}) 
&= \|(I - P)\, g((T_{\mathrm{row}} U T_{\mathrm{col}}^\top + E)x_{\mathrm{OoD}})\| \\
&\ge \big|\|(I - P)\,g(T_{\mathrm{row}} U T_{\mathrm{col}}^\top x_{\mathrm{OoD}})\| \\
&\quad - \|(I-P)[g((T_{\mathrm{row}} U T_{\mathrm{col}}^\top+E)x_{\mathrm{OoD}}) \\
&\qquad - g(T_{\mathrm{row}} U T_{\mathrm{col}}^\top x_{\mathrm{OoD}})]\|\big| \\
&\ge S_{\mathrm{OoD}} - \|I - P\|_{\sigma}\|E x_{\mathrm{OoD}}\|\\
&\ge S_{\mathrm{OoD}} - \|E\|\,(\text{since }\|I - P\|_{\sigma}=1)
\end{aligned}
\end{equation}
Thus, we have:
\begin{equation}
\inf_{x_{\mathrm{OoD}}} d(x_{\mathrm{OoD}}) \ge S_{\mathrm{OoD}} - \|E\|
\end{equation}

\medskip
\noindent\textbf{Step 4: Strict separation condition:}  
Given the empirical condition:
\begin{equation}
S_{\mathrm{OoD}} > 2\,\|E\|
\end{equation}
we directly obtain the strict separation:
\begin{equation}
\begin{aligned}
\sup_{x_{\mathrm{ID}}} d(x_{\mathrm{ID}}) &\le \|E\| < S_{\mathrm{OoD}} - \|E\|\\
&\le \inf_{x_{\mathrm{OoD}}} d(x_{\mathrm{OoD}})
\end{aligned}
\end{equation}

This proves that ID residuals lie strictly below OoD residuals, establishing the empirical spectral residual separation.

\subsection*{Appendix E: Hyperparameter Settings}

The training follows a variational Bayesian approach with inducing point approximations applied to both convolutional and linear layers. Inducing point sizes, prior configurations, and regularization parameters are carefully chosen to balance expressivity and tractability.

The model is trained for 200 epochs using the Adam or SGD optimizer with standard data augmentations and label smoothing. The variational inference is configured using a diagonal Gaussian for the posterior over inducing variables, and KL annealing is employed during early training. Table~\ref{tab:bayesian_hyperparams} lists all key hyperparameters.

\begin{table}[!ht]
\caption{Training Hyperparameters for Bayesian ResNet}
\centering
\renewcommand{\arraystretch}{1.2}
\begin{tabular}{ll}
\midrule
\textbf{Parameter} & \textbf{Value} \\
\midrule
Dataset & CIFAR-10 / CIFAR-100 \\
Input Size & $32 \times 32$ \\
Model & ResNet-18 \\
Epochs & 200 \\
Train Samples & 1 \\
Test Samples & 8 \\
Batch Size (train/test) & 100 / 200 \\
Learning Rate & \(\eta = 10^{-3}\) \\
Optimizer & Adam / SGD \\
Momentum (SGD) & \(\mu = 0.9\) \\
Milestones & [100] \\
Learning Rate Decay & \(\gamma = 0.1\) \\
Seed & 42 \\
Data Augmentation & Crop + Flip \\
Label Smoothing & \(\epsilon = 0.05\) \\
\midrule
\multicolumn{2}{l}{\textbf{Bayesian Inference (Inducing Points)}} \\
\midrule
Inference Type & Inducing Point \\
Conv2d Inducing Size & \(128 \times 128\) \\
Linear Inducing Size & \(100 \times 128\) \\
Whitened Inducing Variables & True \\
$q(\mathbf{u})$ Covariance & Diagonal \\
Learn $\lambda$ & True \\
Initial $\lambda$ & \(\lambda_{\text{init}} = 0.001\) \\
Max $\lambda$ & \(\lambda_{\text{max}} = 0.03\) \\
Prior Standard Deviation & \(\sigma_{\text{prior}} = 1.0\) \\
Max Std. Dev. of $\mathbf{u}$ & \(\sigma_u^{\text{max}} = 0.1\) \\
Cache Cholesky Decomposition & True \\
Sqrt Width Scaling & True \\
Key Layers & \texttt{layer2.1.conv1\&2}, \\
           & \texttt{layer4.1.conv1\&2} \\
\midrule
\end{tabular}
\label{tab:bayesian_hyperparams}
\end{table}

The choice of hyperparameters reflects a balance between modeling flexibility and computational efficiency. In the optimization configuration, we adopt standard values for learning rate and momentum following common deep learning practices, while also introducing a learning rate scheduler to facilitate convergence. Label smoothing is applied with a moderate coefficient \(\epsilon = 0.05\) to mitigate overconfidence in classification outputs.
For the Bayesian inference procedure, we apply variational approximations with inducing point parameterizations. The number of inducing points is selected based on empirical validation: \(128 \times 128\) for convolutional layers and \(100 \times 128\) for fully connected layers, which ensures a good trade-off between approximation accuracy and memory footprint.
We employ diagonal covariance in the variational posterior over the inducing variables \(q(\mathbf{u})\) to reduce computational complexity, along with whitening and Cholesky caching to stabilize training and accelerate matrix operations. The scale parameters \(\lambda\) and \(\sigma\) are initialized conservatively and gradually increased, with learned constraints to avoid numerical instability.

\subsection*{Appendix F: Extra Lemma}

\begin{mybar}
\begin{Lemma}[Conditional-Gaussian Preservation]
\label{thm:gauss-post}
Under the SVGP variational approximation
\begin{equation}
\label{eq:svgp_family}
q(f,u) = p(f \mid u)\,q(u),
\end{equation}
where $p(f \mid u)$ is the prior-conditional Gaussian and $q(u)$ is the variational posterior:

\begin{enumerate}[label=(\alph*)]
\item \textbf{Gaussian $q(u)$.}  
If 
\begin{equation}
\label{eq:qu_gaussian}
q(u) = \mathcal{N}(m,S),
\end{equation}
then the marginal
\begin{equation}
\label{eq:qf_gaussian}
q(f) = \int p(f\mid u)\,q(u)\,du
\end{equation}
is Gaussian with closed-form mean and covariance.

\item \textbf{Flow-based $q(u)$.}  
If $q(u)$ is parameterised by a normalising flow,
\begin{equation}
\label{eq:qu_flow}
q(u) = g_{\phi}\,\#\,q_{0}, \quad q_{0}=\mathcal{N}(m,S),
\end{equation}
then $q(f)$ is in general non-Gaussian (a mixture). However,
$p(f\mid u)$ remains Gaussian for any $u$, and expectations under $q(f)$ are computed by reparameterised Monte Carlo.
\end{enumerate}
\end{Lemma}
\end{mybar}

\noindent\textbf{Proof Sketch.} The conditional $p(f\mid u)$ is linear-Gaussian. Convolving it with a Gaussian $q(u)$ yields a Gaussian marginal (case (a)). With a flow-based, non-Gaussian $q(u)$, the marginal becomes a mixture (case (b)); the conditional form is preserved, and expectations can be estimated via sampling.